\documentclass[sigconf]{acmart}
\AtBeginDocument{%
  }

\setcopyright{acmlicensed}
\copyrightyear{2026}
\acmYear{2026}
\acmDOI{XXXXXXX.XXXXXXX}
\acmConference[Conference acronym 'XX]{Make sure to enter the correct
  conference title from your rights confirmation email}{June 03--05,
  2018}{Woodstock, NY}
\acmISBN{978-1-4503-XXXX-X/2018/06}

\usepackage{graphicx}
\usepackage{algorithm}
\usepackage{algorithmic}
\usepackage{booktabs}
\usepackage[table]{xcolor}
\usepackage{tabularx}
\usepackage{booktabs}
\usepackage{multirow}
\usepackage{amsmath}
\usepackage{amsthm}
\usepackage{makecell}
\usepackage{fontawesome}
\newcommand{\tokred}[1]{\,{\scriptsize\color{green!45!black}($\downarrow$#1\%)}}

\begin{document}

\title{Learning What to Remember and What to Internalize in LLM\\ Self-Evolution via Adaptive Memory-Parameter Coordination}

\author{Tianyun Ji}
\email{jitianyun2002@mail.ustc.edu.cn}
\orcid{0009-0000-9490-1659}
\affiliation{%
  \institution{University of Science and Technology of China}
  \city{Hefei}
  \state{Anhui}
  \country{China}
}

\author{Zhenya Huang}
\correspondingauthor
\email{huangzhy@ustc.edu.cn}
\orcid{0000-0003-1661-0420}
\affiliation{%
  \institution{University of Science and Technology of China}
  \city{Hefei}
  \state{Anhui}
  \country{China}
}

\author{Jiayu Liu}
\email{jiayuliu@cityu.edu.hk}
\affiliation{%
  \institution{City University of Hong Kong}
  \city{Hong Kong}
  \country{China}
}

\author{Zirui Liu}
\orcid{0009-0002-7263-9607}
\affiliation{%
  \institution{University of Science and Technology of China}
  \city{Hefei}
  \state{Anhui}
  \country{China}
}

\author{Yu Su}
\email{yusu@hfnu.edu.cn}
\affiliation{%
  \institution{Hefei Normal University}
  \city{Hefei}
  \state{Anhui}
  \country{China}
}

\author{Hongbin Pei}
\email{peihongbin@xjtu.edu.cn}
\affiliation{%
  \institution{Xi'an Jiaotong University}
  \city{Xi'an}
  \state{Shanxi}
  \country{China}
}

\renewcommand{\shortauthors}{Ji et al.}

\begin{abstract}
Large language model agents increasingly operate in dynamic environments where tool interfaces, APIs, and user requirements change after deployment. Existing self-evolution methods mainly follow two paradigms: harness-based approaches, which externalize feedback into editable memories or skills for rapid adaptation, and parameter-based approaches, which internalize experience into model parameters for deeper capability improvement. However, using either mechanism alone creates a trade-off between flexibility and performance. This paper asks how an agent can coordinate both channels to achieve robust self-evolution. We present COVE, a unified agent self-evolution framework that combines harness-based and parameter-based learning through task-aware routing, stage-aware scheduling, and knowledge optimization. Through this design, COVE treats self-evolution not as indiscriminate accumulation of experience, but as a coordinated process that matches tasks and knowledge types to appropriate learning mechanisms. Experiments across multiple task categories show that COVE outperforms single-channel evolution strategies, demonstrating more robust and efficient improvement under changing environments.
\end{abstract}



\received{26 July 2026}

\maketitle

\section{Introduction}
Large language model (LLM) agents have rapidly moved from passive text generators to interactive systems that can plan, call tools, write code, operate in embodied environments, and solve specialized tasks through multi-step reasoning~\cite{yao2022react,schick2023toolformer,gao2025agent4edu,zhan2025coderagent,yu2025gar}.

This dependency creates a fundamental tension. The external world evolves continuously. APIs are updated, database schemas change, new libraries appear, domain conventions shift, and users introduce tasks that were not anticipated during development\cite{ning2025survey, zhang2024multimodal}. Maintaining a high-performing agent under such conditions requires repeated reconfiguration or retraining, which is expensive in both human labor and computation. 

\begin{figure*}[t]
  \centering
  \includegraphics[width=1\textwidth]{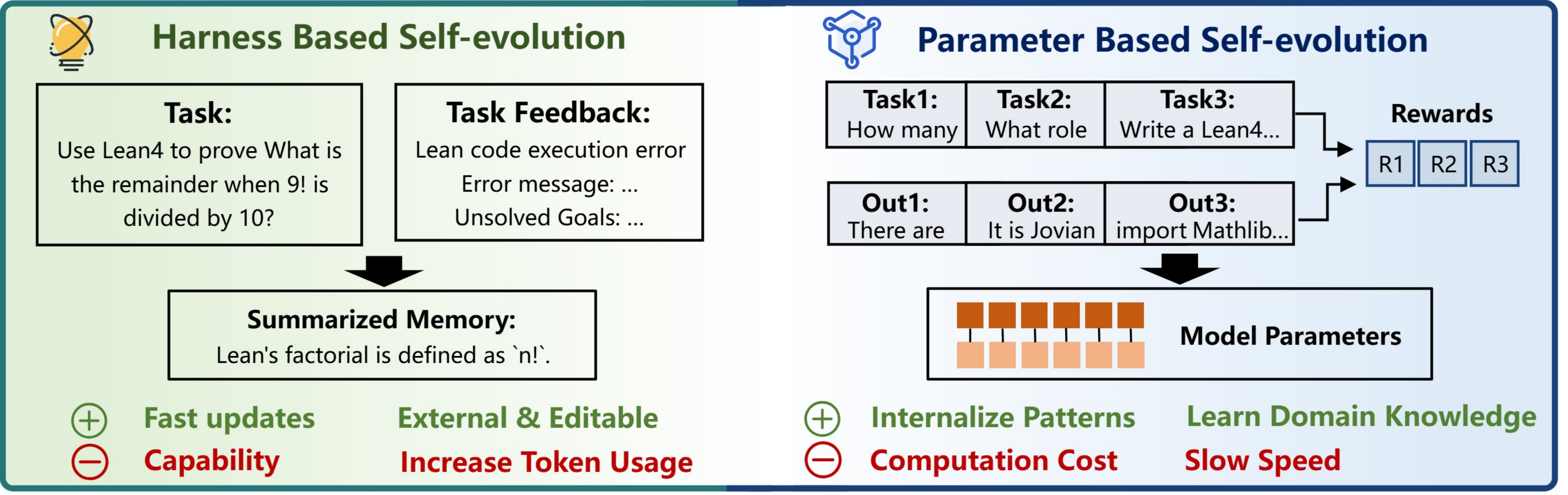}
  \caption{Harness-based evolution keeps editable memories or skills, while parameter-based evolution updates model weights.}
  \label{fig:1_PGchannel}
  \vspace{-0.3cm}
\end{figure*}

Agent self-evolution offers a promising path. Instead of treating deployment as the end of learning, a self-evolving agent treats task execution as a source of supervision~\cite{fang2025comprehensive}. By converting environment feedback into reusable experience, the agent can keep improving on new tasks, gradually learning new tools, adapting to changing environments, and acquiring specialized domain knowledge from interaction~\cite{hu2025agentgen}.
As shown in Figure~\ref{fig:1_PGchannel}, existing self-evolution methods can be broadly grouped into two paradigms. \textbf{Harness-based} methods, represented by systems such as Evo-Memory~\cite{wei2025evo}, summarize feedback into external memories, skills that are retrieved and injected into the agent harness at inference time. This paradigm is fast, and well suited to changing task surfaces, because new experience can be easily added or revised. Yet it can produce misleading or unusable experience when the base model lacks the competence to interpret feedback, as observed in prior work on self-generated skills~\cite{huang2024large} and embodied curricula~\cite{wang2023voyager}.
\textbf{Parameter-based} methods take the complementary route. Works such as WizardLM~\cite{xu2024wizardlm} and Self-Challenging Agent~\cite{zhou2026self} use generated data, task rewards, or feedback to update model weights through supervised fine-tuning or reinforcement learning. Such updates can internalize deeper patterns, including latent strategies or domain-specific reasoning routines that cannot be reliably captured by a short memory entry. However, parameter-based learning is costly and slow to adapt, which is unnecessary for routine updates that a small editable memory can handle. 

The limitations of the two paradigms are complementary, but each limitation is serious. Effective self-evolution therefore requires coordination between the two channels rather than reliance on either one alone. To achieve this, we conduct an analysis of the agent's evolution process. On Lean theorem-proving tasks from MiniF2F~\cite{zheng2021minif2f}, harness-side memory accumulation improves success by less than 3\%, even when the number of retrieved memories per task increases from 2 to 8. At the intra-task level, plateau-triggered parametric updates outperform both always-on parametric training and harness-only evolution in the stage-aligned comparison, showing that the timing of training is itself a key decision. This suggests that the choice of evolution channel must be conditioned jointly on the task category and learning stage.

Second, self-evolution must distinguish among different knowledge types. In the API-renaming analysis on WikiTableQuestions~\cite{pasupat2015compositional}, the API-call correctness of direct parametric learning model falls from 96.50\% to 54.00\% after interface names change, which indicates that this task contains heterogeneous knowledge that should not be stored in a single way. We identify \textbf{volatility} as the key criterion for deciding whether knowledge should be externalized or internalized: volatile surface forms such as library-version details should remain in the harness. Stable principles such as reusable proof tactics are better candidates for parametric internalization.

Based on these observations, we propose \textbf{COVE} (Channel Orchestrated Volatility-aware Evolution), an agent self-evolution framework that coordinates harness-based and parameter-based learning under a unified pipeline. COVE is built around three operational mechanisms. First, a \textbf{Task-aware Router} makes feedback-conditioned channel decisions, assigning each task to a specific channel according to task characteristics, execution feedback, and observed failure signals. Second, a \textbf{Stage-aware Scheduler} turns channel switching into a measurable trigger problem. It monitors plateau, data sufficiency, and cold-start failure to decide when to continue harness-based exploration and when to initiate parametric training. Third, \textbf{KnowledgePO} performs dual-modal knowledge optimization. Harness-side memories improve rollout and data collection, while parametric-side learning selectively internalizes stable knowledge, and uses anti-recitation signals to prevent volatile knowledge from being memorized into parameters.

In summary, this paper makes the following contributions:
\begin{itemize}
  \item We provide a systematic analysis of agent self-evolution from the perspectives of task characteristics, learning stages, and knowledge volatility, clarifying when harness-based and parameter-based methods should be used.
  \item We introduce COVE, a self-evolution framework that integrates task routing, intra-task scheduling, and dual-modal knowledge optimization to combine the flexibility of memory with the depth of parameter learning.
  \item We empirically show that the proposed framework outperforms single-channel evolution strategies across different tasks, demonstrating that coordinated evolution yields more robust and efficient improvement than either harness-based or parameter-based learning alone.
\end{itemize}

\section{Related Work}
Current approaches to agent self-evolution can be broadly organized into two paradigms. They differ fundamentally in where the acquired experience is stored and how it is reused.


\subsection{Harness-Based Methods}
Harness-based self-evolution improves an agent by augmenting the information available in its execution harness. In this paradigm, an LLM is coupled with prompts, tools, retrieval modules, and external storage~\cite{zhai2025agentevolver}. After each interaction, the agent distills environmental feedback into reusable artifacts, such as memories or executable skills. Reflexion~\cite{shinn2023reflexion} uses verbal feedback as reusable experience for later trials. Voyager~\cite{wang2023voyager}, for example, demonstrates how an embodied agent can continually acquire new skills from open-ended exploration. Evo-Memory~\cite{wei2025evo} further studies test-time learning through self-evolving memory, where accumulated experience is retrieved to guide future decisions. In these systems, the agent appears to ``evolve'' because its accessible knowledge base expands. Yet harness-based evolution can produce misleading experience when the base model lacks the competence to interpret feedback.

\subsection{Parametric-Based Methods}
Parameter-based self-evolution takes a complementary approach. It internalizes environmental signals directly into the model parameters through supervised fine-tuning (SFT) or reinforcement learning (RL)~\cite{schulman2017proximal, sun2025seagent, lin2018efficient, bailey2026scaling}. WizardLM~\cite{xu2024wizardlm} shows that LLMs can bootstrap stronger instruction-following behavior by generating complex synthetic instructions and fine-tuning on the resulting data. More recent work, such as Self-Challenging Agent~\cite{zhou2026self}, uses task outcomes as reward signals and updates the agent policy through reinforcement learning. In this paradigm, evolution is reflected in the model itself. The model parameters change, and the agent can acquire more stable reasoning routines that no longer depend on explicitly memory retrieving. Yet such adaptation requires intensive training, making it unnecessary for routine updates that could instead be handled by a small editable memory.
\section{Evolving Paradigm Analysis}
\label{sec:Analysis}

In this section, we analyze the agent self-evolution process and show that both harness-based and parameter-based self-evolution have their own limitations. Effective self-evolution therefore requires coordination between these two channels.

\subsection{Task-Dimension Analysis}
We first test whether the two channels are interchangeable across tasks by designing two diagnostic tasks with different structural requirements. The first task is Lean theorem proving, where trajectories are collected from MiniF2F~\cite{zheng2021minif2f} and harness-based self-evolution is applied to test whether accumulated harness-side knowledge can improve formal proof construction.

\begin{figure}
  \centering
  \includegraphics[width=0.5\textwidth]{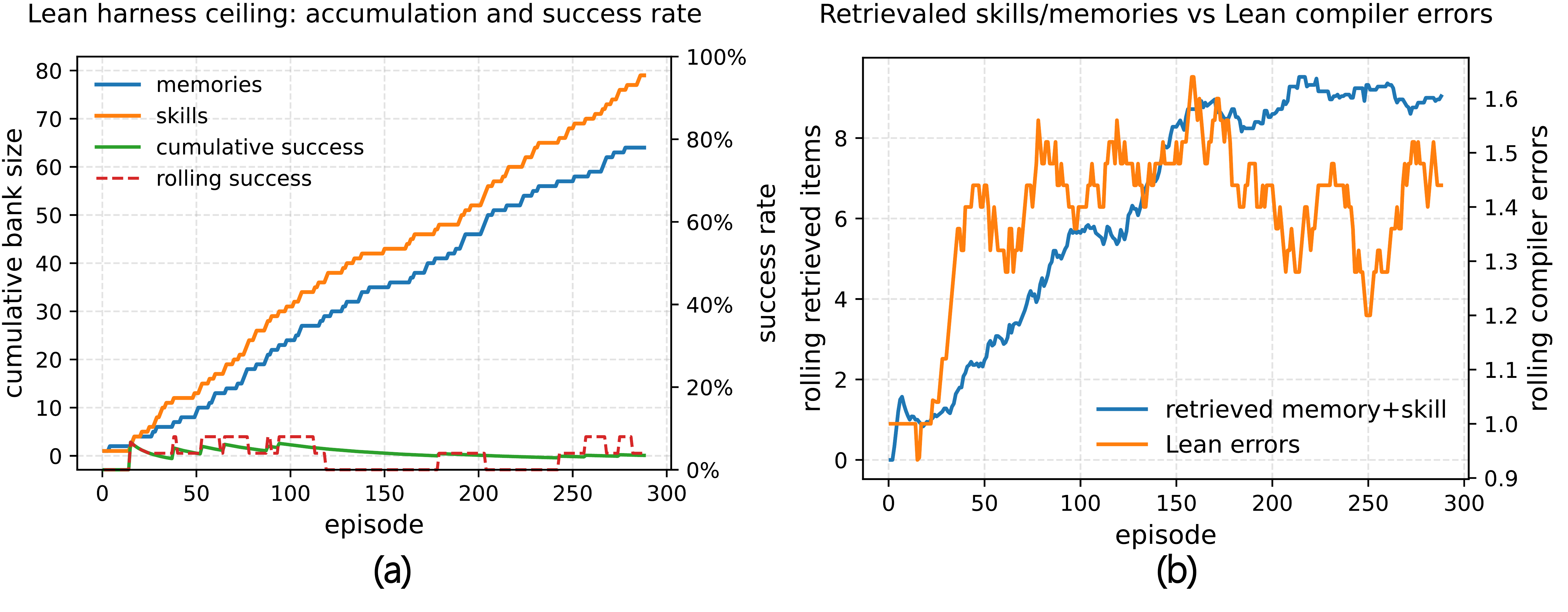}
  \caption{Lean harness evolution: more stored or retrieved knowledge yields little success gain or error reduction.}
  \label{fig.3_prompt_failure}
\end{figure}

\begin{figure}
  \centering
  \includegraphics[width=0.5\textwidth]{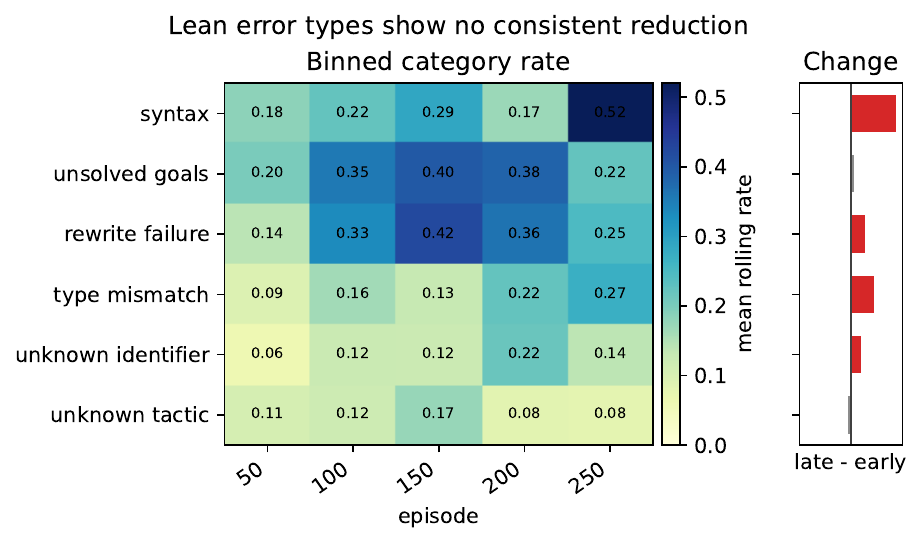}
  \caption{Lean error types under harness evolution, showing no consistent reduction across episodes.}
  \label{fig.3_lean_error_types}
\end{figure}

The second task is table question answering with dynamic tool interfaces, where trajectories are collected from WikiTableQuestions~\cite{pasupat2015compositional} and used to train a parameter-based self-evolution model based on Qwen3-8B~\cite{yang2025qwen3}. We then evaluate the evolved model under two conditions: one keeps the API names unchanged from training, and the other manually renames the APIs at test time. 

 \textbf{At the inter-task level}, the two tasks exhibit different failure patterns. In the Lean theorem-proving task, Figure~\ref{fig.3_prompt_failure}(a) shows that the agent accumulates more knowledge across episodes, but success rate remains nearly flat. Figure~\ref{fig.3_prompt_failure}(b) further shows that retrieving more Memory/Skill entries does not correspond to fewer Lean errors, and Figure~\ref{fig.3_lean_error_types} shows that proof-error categories do not consistently decline. As episodes increase from 50 to 250, compiler errors do not decline with self-evolution. In fact, the most basic syntax errors even become more frequent. Harness memory can store hints or tactic patterns, but it cannot reliably create the operational competence needed to execute them.
In the dynamic-tool task, parametric evolution performs much better when the API names remain unchanged, but its performance drops sharply after renaming. As shown in Table~\ref{tab:gradient-api-renaming}, success rate falls from 40.5\% to 16.5\%, and API-call correctness falls from 96.5\% to 54.0\%. This suggests that parametric evolution can make the model brittle when the surface interface changes. 

These cross-task results reveal complementary bottlenecks. Volatile tasks require editable harness-side adaptation, while competence-limited tasks require parametric-side learning. We therefore separate channel selection into two levels: \textbf{ an \emph{inter-task} decision about which channel fits a task family, and an \emph{intra-task} decision about when the channel should change during learning}.


\begin{figure}
  \centering
  \includegraphics[width=0.47\textwidth]{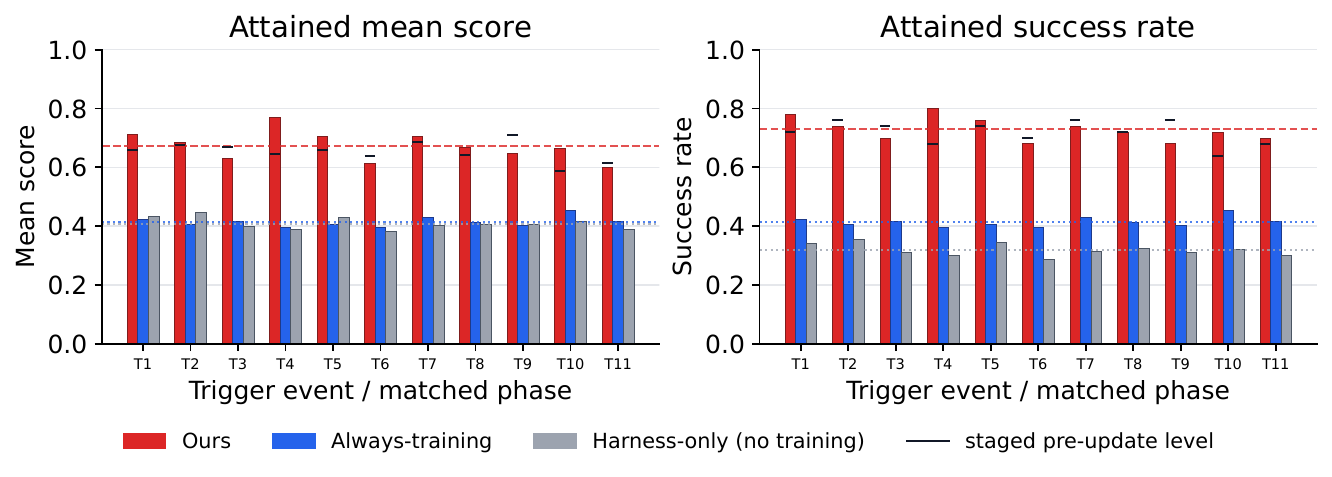}
  \caption{Stage-aligned comparison: plateau-triggered updates outperform always-on parametric and harness-only baselines.}
  \label{fig.trigger_before_after_deltas}
\end{figure}

\textbf{At the intra-task level}, the preferred channel may also change over time. A common path is harness-first exploration: the agent stores feedback and accumulates high-quality trajectories. Once performance plateaus or sufficient data are available, parametric training can consolidate the stable part of this experience. Figure~\ref{fig.trigger_before_after_deltas} illustrates why the timing of this transition matters. When the three mechanisms are aligned by training stages, plateau-triggered parametric updates reach higher attained mean score and success rate than both continuous parametric training and harness-only evolution, which remain roughly flat. This suggests that the benefit does not come merely from using parameter updates, but from applying them when harness-side exploration has exposed a stable bottleneck and accumulated useful signal. The reverse path is also possible. In cold-start domains such as Lean theorem proving, the base model may fail to produce useful trajectories, so an initial parametric update may be needed before harness-side evolution becomes productive. Thus, intra-task coordination should be driven by success rate, plateau signals, data sufficiency, and cold-start failure rather than by a fixed schedule.

\subsection{Knowledge-Dimension Analysis}

\begin{table}[t]
\centering
\small
\caption{Parameter-based self-evolution under API changes.}
\label{tab:gradient-api-renaming}
\begin{tabular}{p{0.4\linewidth} p{0.2\linewidth} p{0.28\linewidth}}
\hline
\textbf{Test Setting} & \textbf{Success Rate} & \textbf{API Correct Rate} \\
\hline
Same API names & 40.5\% & 96.5\% \\
Renamed API names & 16.5\% & 54.0\% \\
Volatility-aware on Renamed & 32.4\% & 92.5\% \\
\hline
\end{tabular}
\end{table}
The task-level analysis determines which evolution channel should be activated, but it does not yet determine what should be stored in each channel. This matters because a collaborative agent may train on trajectories produced with memories, storing knowledge implicitly in weights. The central question is therefore: \textbf{which parts of those memories should be allowed to become parameters?}

A tempting criterion is frequency. If a piece of knowledge appears repeatedly and helps the agent succeed, it may seem valuable to absorb it into the model parameters. However, frequency alone is an unsafe signal, because it conflates usefulness with durability. A counterintuitive but important case is high-frequency interface knowledge, which can be among the most dangerous forms of knowledge to internalize. Table~\ref{tab:gradient-api-renaming} makes this concrete: API calls constitute highly frequent and practically useful knowledge, yet directly internalizing such surface-level API knowledge makes the model brittle under version changes, leading to a 59.3\% drop in success rate in the second row. The problem is therefore not that either self-evolving channel is inherently flawed, but that different types of knowledge should be processed through different mechanisms.
\begin{itemize}
  \item \textbf{Volatile knowledge} depends on external surface forms, such as API names or table schemas. It should remain in Memory and be explicitly protected from internalization during training.
  \item \textbf{Stable knowledge} captures durable underlying patterns, such as algorithmic idioms or domain reasoning routines. It is a suitable candidate for parameter-based internalization.
  \item \textbf{Strategic knowledge} consists of higher-level heuristics, such as debugging plans. It should remain in Memory and be internalized only after repeated evidence of stability.
\end{itemize}
As shown in Table~\ref{tab:gradient-api-renaming}, after annotating knowledge volatility and introducing corresponding rewards, the model still achieves relatively stable performance, with a 92.5\% API-call success rate on renamed tasks.

In summary, effective self-evolution requires deciding which evolution channel should process each experience, when the channel should change, and where the resulting knowledge should reside. The following method section operationalizes these decisions as task-aware routing, stage-aware scheduling, and KnowledgePO.

\section{Method}

\begin{figure*}
  \centering
  \includegraphics[width=1\textwidth]{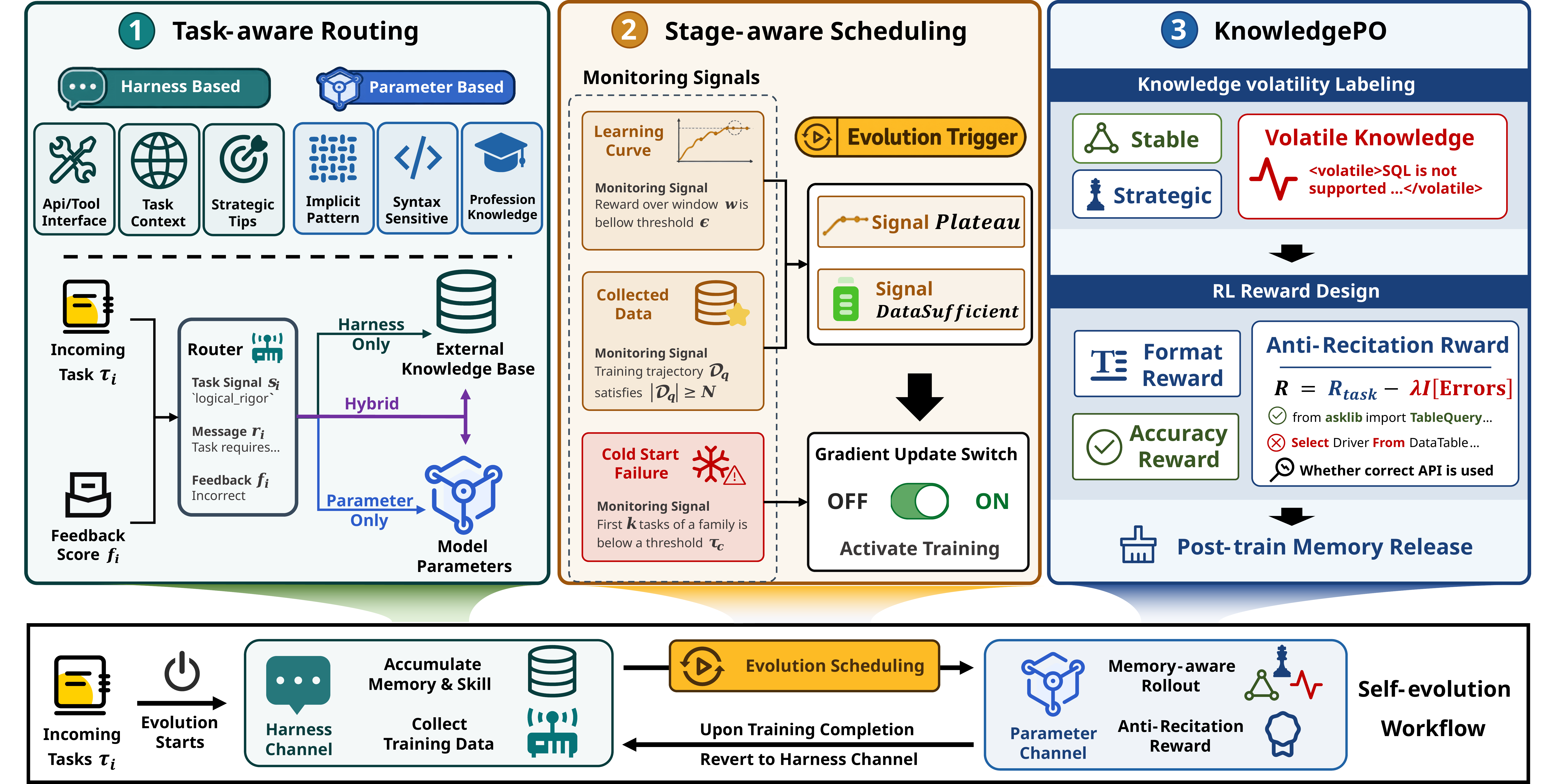}
  \caption{Overview of COVE. Given task feedback, the task-aware router selects an evolution channel, the stage-aware scheduler determines when to invoke harness-side or parameter-side updates, and KnowledgePO coordinates knowledge transfer between external memories and model parameters by preserving volatile knowledge in the harness.}
  \label{fig.4_framework}
\end{figure*}

Based on these observations, we propose \textbf{COVE} (Figure~\ref{fig.4_framework}), a cooperative self-evolution framework that integrates harness-based adaptation with parameter-based optimization. The central premise of COVE is that self-evolution should not be treated as a monolithic operation. Different tasks expose different kinds of learnable structure. COVE therefore assigns each task to the evolution channel that best matches its knowledge structure, and further allows the two channels to reinforce each other.


\subsection{Task Formulation}

We consider an agent operating over a sequential stream of tasks $\{\tau_i\}_{i=1}^{T}$, where each task is sampled from a task distribution or task family $\mathcal{D}$. The agent is composed of a parametric policy model $\pi_{\theta}$, an external knowledge base $\mathcal{K}=\{k_j\}$, and an execution environment $\mathcal{E}$. After attempting task $\tau_i$, the environment returns feedback $f_i$, which may include task rewards, success or failure labels, error traces, tool-call logs, and automatic or human evaluation results.

COVE supports two classes of updates. The harness-side update,
\[
\mathcal{K} \leftarrow U_h(\mathcal{K}, \tau_i, f_i),
\]
extracts reusable memory or skill entries from feedback and updates the external knowledge base. The parametric-side update,
\[
\theta \leftarrow U_{\theta}(\theta, \mathcal{D}_{train}, \mathcal{K}),
\]
optimizes the parameters of the base policy using accumulated  trajectories and knowledge-augmented training data. The objective is not to maximize the size of $\mathcal{K}$, but to maximize cumulative task utility under both inference-time and training-time budgets.
\[
\max \sum_{i=1}^{T} R_i
- \alpha \sum_{i=1}^{T} C^{\text{tok}}_i
- \beta \sum_{m=1}^{M} C^{\text{train}}_m ,
\]
where $R_i$ denotes task reward, $C^{\text{tok}}_i$ denotes the prompting token cost, and $C^{\text{train}}_m$ denotes the compute cost of $m$-th parametric update.
\subsection{Task-aware Routing}
\label{sec:task-aware-routing}
Applying harness-based and parameter-based self-evolution to every task simultaneously is neither economical nor robust.
COVE therefore routes each task according to the kind of knowledge it demands and the type of failure it exposes.

We instantiate the router as a \textit{feedback-conditioned decision policy} rather than a hand-tuned scoring function. Given a task description $\tau_i$ and the execution feedback $f_i$---a compact record comprising an outcome status, a scalar score, and a diagnostic message emitted by the evaluator---the router queries a constrained judge that returns a structured decision record \(d_i = (m_i,\ s_i,\ r_i)\), where $m_i$ denotes the selected evolution channel, $s_i$ denotes a set of short free-form descriptors characterizing why the task behaves as it does (for instance \texttt{dynamic\_tool} or \texttt{logical\_rigor}), and $r_i$ denotes a one-sentence justification. Rather than composing the decision from a fixed-dimensional signal vector, the judge emits the channel directly and attaches the signals as an interpretable trace, which keeps the routing decision inspectable.

The channel is drawn from three options, denoted by $\mathcal{M}=\{\texttt{harness\_only},\texttt{parametric\_candidate},\texttt{hybrid}\}$. \texttt{harness\_only} is selected when failures stem from volatile or in-context knowledge that does not benefit from weight updates. \texttt{parametric\_candidate} is selected when the task rewards an internalizable competence that is hard to express as a prompt. \texttt{hybrid} is reserved for cases exhibiting a clear signal of each kind, for example a formal-reasoning task that also requires tool or retrieval interaction.

\subsection{Stage-aware Scheduling}
Beyond routing by task type, it is equally important to determine when each evolution channel should be invoked, since the two channels exhibit stage-dependent trade-offs. Harness-based evolution is fast, but may saturate after early gains. While parameter-based evolution can offer deeper improvement, it incurs substantially higher computational cost and latency. One way is to begin with harness-side exploration and switch to parametric-side training once performance plateaus or sufficient trajectories have been collected. This strategy provides the agent with a low-cost entry point into the task distribution and allows it to accumulate memories that can benefit subsequent training. While in regimes such as Lean theorem proving, the base model may fail to generate useful trajectories or memories without first acquiring basic syntactic and procedural competence. In such cases, a preliminary parametric update may be necessary to activate the behaviors required for effective harness-side self-evolution.

Therefore, COVE does not fix the order or frequency of the two evolution channels. Instead, it introduces a \textit{Stage-aware Evolution Trigger}, which dynamically decides when to invoke each channel based on measurable learning signals, denoted by $\textsc{Trigger}(\cdot)$, that continuously monitors the current task family and decides whether to continue harness-side exploration or initiate a parametric update. The trigger is based on three reproducible conditions. \textbf{Performance plateau}, denoted by $\textsc{Plateau}$, occurs when the improvement in success rate or average reward over a recent window of $w$ episodes is below a threshold $\epsilon$. \textbf{Data sufficiency}, denoted by $\textsc{DataSufficient}$, holds when the number of high-quality trajectories $\mathcal{D}_q$ satisfies $|\mathcal{D}_q| \geq N$. \textbf{Cold-start failure}, denoted by $\textsc{ColdStartFailure}$, holds when the success rate in the first $k$ tasks of a family is below a threshold $\tau_c$, indicating that harness-side evolution cannot bootstrap itself.

For a task family routed as \texttt{parametric\_candidate} or \texttt{hybrid}, COVE triggers a parametric update when
\[
\textsc{Trigger} =
\textsc{Plateau}
\lor
\textsc{DataSufficient}
\lor
\textsc{ColdStartFailure}.
\]
The first two conditions support the common path of harness-first exploration followed by parametric-side internalization. The third condition captures the opposite path, where parametric training must first create a minimal competence floor. This trigger also yields a clean experimental interface. It can be ablated against no-trigger and fixed-interval schedules while keeping the router and training pipeline unchanged.

\subsection{KnowledgePO: Dual-modal Knowledge Optimization}

Routing and scheduling determine when each channel is used. KnowledgePO determines how the channels exchange knowledge. Its goal is to turn harness-based and parameter-based self-evolution into a closed loop: external memories improve the quality of training rollouts, and trained parameters reduce future dependence on external memories when the relevant knowledge has become stable and internalized. This dual-modal optimization is essential because a knowledge base that only grows will eventually increase context cost, whereas a model that indiscriminately internalizes all retrieved knowledge will memorize obsolete surface forms.

\subsubsection{Harness-assisted Parametric: Memory-aware Rollout}
Before training starts, COVE assigns a volatility label $\ell_j$ to each knowledge entry $k_j$ when the entry is created. After a task is completed, the harness-side module summarizes the task content, execution trace, model behavior, and feedback into candidate memory or skill entries, denoted by $\mathcal{C}_i$. For each candidate $c \in \mathcal{C}_i$, a constrained judge emits a label $\ell(c) \in \mathcal{L}$, where $\mathcal{L}=\{\texttt{volatile}, \texttt{stable}, \texttt{strategic}\}$, using the criteria described in Section~\ref{sec:Analysis}.

Moreover, the label is not treated as a one-shot annotation. When an initially non-volatile entry is repeatedly revised, the system treats repeated correction as evidence that the entry depends on shifting external state and promotes it to \texttt{volatile}. During retrieval and rollout collection, entries labeled or promoted as volatile are wrapped with explicit markers such as \texttt{<volatile>...</volatile>}.

With this collected knowledge, COVE performs \textit{memory-aware inference} during rollout. Before each attempt, the agent retrieves the top-$k$ memory or skill entries relevant to the task, denoted by $\mathcal{K}_r \subseteq \mathcal{K}$, and injects them into the message list. This allows the model to exploit knowledge discovered during previous harness-side exploration, increasing rollout success rate and improving the quality of trajectories $\mathcal{D}_{rollout}$ used for supervised or reinforcement learning updates. In effect, the knowledge base acts as a temporary scaffold that raises the density of useful training signal.

For volatile entries, COVE applies an \textit{anti-recitation reward}. Before training starts, names inside volatile entries are randomly renamed. Let $\mathcal{N}_i$ denote the legal volatile names provided by the currently retrieved memory in episode $i$. The only legal names in the current episode are those in $\mathcal{N}_i$, not names that appeared in historical training contexts. If the model calls an obsolete or unobserved volatile name, it reveals reliance on memorization rather than retrieval-conditioned execution and receives a penalty:
\[
R = R_{\text{task}} -
\lambda \cdot
\mathbb{I}[\text{uses stale or unobserved volatile name}].
\]
Here $R_{\text{task}}$ measures task correctness and format validity, while the penalty discourages shortcuts from fixed interface names to fixed actions. The resulting behavior is not ``remember and repeat'', but ``read the current entry and act accordingly''. Stable knowledge is exempt from this penalty, allowing reusable patterns to be internalized into the model weights.

\subsubsection{Parametric-assisted Harness: Memory Release}

Since accumulated memories consume context and increase retrieval noise, COVE evaluates whether stable knowledge still needs to remain in the online retrieval set after training, denoted by $\mathcal{K}_{online}$.  For each stable  memory $k_j$, COVE runs an A/B evaluation on a held-out subset $\mathcal{D}_{held}$. Condition A includes $k_j$ in the retrieved context, while condition B removes it. If the performance drop
\[
\Delta_j = \text{Score}_{A}(k_j) - \text{Score}_{B}(\varnothing)
\]
is below a threshold after training, and if $k_j$ previously had positive utility before training, the entry is marked as \texttt{internalized} and released from online retrieval. The entry may remain in an archival store for auditability, but it will no longer be retrieved. Volatile memories are never released, since their value lies in carrying current external symbols rather than in encoding stable competence.

This distinction keeps the knowledge base compact without erasing knowledge that must remain external. KnowledgePO therefore maintains a high-quality retrieval space while allowing the model parameters to absorb only the parts of experience that are durable enough to be worth internalizing. A procedural summary of the full COVE workflow is provided in Appendix~\ref{app:cove-workflow}. The framework therefore avoids both extremes. It neither relies indefinitely on an ever-growing harness memory, nor forces every piece of experience into model parameters regardless of volatility. Our code is available at \url{https://anonymous.4open.science/r/cove-8BCC/}.

\section{Experiments}
\label{sec:experiments}
\begin{table*}[t]
\centering
\caption{Main results: success rate (Succ.) and training token cost (Tok.), with the macro-average cost over the costed tasks (Avg.\ Tok.). Green annotations ($\downarrow$) give the token reduction of Ours relative to Parametric-only. ``--'' marks entries without parametric-side training. The accounting protocol is in Appendix~\ref{app:token-accounting}. Best per column in \textbf{bold}, second best \underline{underlined}.}
\label{tab:exp1_success_rate_cost}
\setlength{\tabcolsep}{4pt}
\begin{tabular}{l c cc cc cc cc c c}
\toprule
\multirow{2}{*}{Method} & Lean4 & \multicolumn{2}{c}{APPS} & \multicolumn{2}{c}{TableQA} & \multicolumn{2}{c}{HotpotQA} & \multicolumn{2}{c}{MATH} & Hybrid & \multirow{2}{*}{Avg.\ Tok.} \\
\cmidrule(lr){2-2} \cmidrule(lr){3-4} \cmidrule(lr){5-6} \cmidrule(lr){7-8} \cmidrule(lr){9-10} \cmidrule(lr){11-11}
 & Succ. & Succ. & Tok. & Succ. & Tok. & Succ. & Tok. & Succ. & Tok. & Succ. & \\
\midrule
Base & 0.0 & 23.2 & -- & 34.3 & -- & 45.1 & -- & 75.1 & -- & 15.6 & -- \\
Evo-Memory & 0.0 & 11.5 & -- & 36.2 & -- & 64.5 & -- & 69.0 & -- & 18.2 & -- \\
Self-Challenging & 3.0 & 26.3 & 28.3K & 27.1 & 28.3K & 57.4 & 28.3K & 72.6 & 28.3K & 14.6 & 28.3K \\
\midrule
Harness-only & 3.0 & 31.6 & -- & \textbf{53.5} & -- & 66.2 & -- & 84.0 & -- & 20.8 & -- \\
Parametric-only & \underline{6.2} & \underline{33.1} & 44.0K & 47.0 & 5.5K & \underline{69.4} & 2.0K & \underline{91.6} & 4.8K & \underline{21.3} & 14.1K \\
\rowcolor{gray!12}
Ours & \textbf{7.0} & \textbf{33.4} & \textbf{5.5K}\tokred{88} & \underline{50.0} & \textbf{1.5K}\tokred{73} & \textbf{69.6} & \textbf{312}\tokred{84} & \textbf{91.7} & \textbf{503}\tokred{90} & \textbf{24.1} & \textbf{2.0K}\tokred{86} \\
\bottomrule
\end{tabular}
\end{table*}

\begin{table}[t]
\centering
\caption{Routing ablation over APPS, MATH, TableQA, and HotpotQA. Token/Inst.: parametric training tokens per routed instance; Rel.: cost relative to Always-Both; Eff.: performance per token (higher is better).}
\label{tab:router_analysis}
\setlength{\tabcolsep}{5pt}
\begin{tabular}{lccc>{\columncolor{gray!12}}c}
\toprule
Method & Succ. & Token/Inst. & Rel. & Eff.\ $\uparrow$ \\
\midrule
Always-Both & \textbf{69.0} & 14.1K & 1.00$\times$ & 4.89 \\
Random-Route & 62.7 & 7.1K & 0.50$\times$ & 8.82 \\
Ours & 65.0 & \textbf{2.3K} & \textbf{0.16}$\times$ & \textbf{28.83} \\
\bottomrule
\end{tabular}
\end{table}

\subsection{Experiment Setup}
We use Qwen3-8B~\cite{yang2025qwen3} as the base model for all experiments and train on 4 NVIDIA A100 80GB GPUs. Unless otherwise specified, the maximum model context length is 16,384 tokens, and the maximum generation length is 8,192 tokens.

We evaluate on five tasks that cover knowledge-intensive question answering, tool-use reasoning, code generation, formal theorem proving, and mathematical reasoning: WikiTableQuestions (TableQA)~\cite{pasupat2015compositional}, HotpotQA~\cite{yang2018hotpotqa}, APPS~\cite{hendrycks2021apps}, MiniF2F~\cite{zheng2021minif2f}, and MATH~\cite{hendrycks2021math}.

We compare against three baselines. Original Qwen3-8B model without self-evolution, Evo-Memory~\cite{wei2025evo} for harness-based self-evolution and Self-Challenging~\cite{zhou2026self} for parameter-based self-evolution. Since Self-Challenging does not provide an official implementation, we implement a reproduction following its reported training protocol. We also include two ablation variants of our method. \textbf{Harness-only} runs only the harness-side evolution component of COVE and does not perform training. \textbf{Parametric-only} treats all collected samples as training data and updates the model without using external Memory or Skill modules.

In Table~\ref{tab:exp1_success_rate_cost}, we report the \emph{success rate} (Succ.), namely the fraction of evaluation instances solved correctly, together with the per-instance training token cost (Tok.) for parametric-side updates. Entries marked ``--'' do not have a comparable parametric-side training cost. Self-Challenging has the same token cost across tasks because it trains one policy model whose cost is amortized uniformly across evaluation tasks. We further explain why and describe the full token accounting protocol in Appendix~\ref{app:token-accounting}.

\subsection{Main Results}
Compared with the base model and the two external self-evolution baselines, COVE achieve better overall performance. Against the two single-channel variants, COVE achieves the best on Lean4, APPS, HotpotQA, and MATH, and remains competitive on TableQA. Importantly, these gains are obtained with 86\% fewer training tokens than indiscriminate parametric learning. These results indicate broader robustness rather than uniform dominance: coordinated self-evolution is most useful when the task benefits from both reusable parametric competence and selectively retrieved external experience.

The last column further evaluates this intended use case. We construct a hybrid test subset by selecting instances that the router classifies as hybrid, namely examples that are high quality and non-trivial. On this subset, COVE obtains the highest success rate, 24.1\%, outperforming both the base model (15.6\%) and the strongest single-channel variant, Parametric-only (21.3\%). This result suggests that the joint harness-plus-parametric training is especially beneficial on the samples for which the router activates both channels, rather than merely improving the average score through task-level effects.

\subsection{Task-aware Router Analysis}

\begin{figure}[t]
  \centering
  \includegraphics[width=0.5\textwidth]{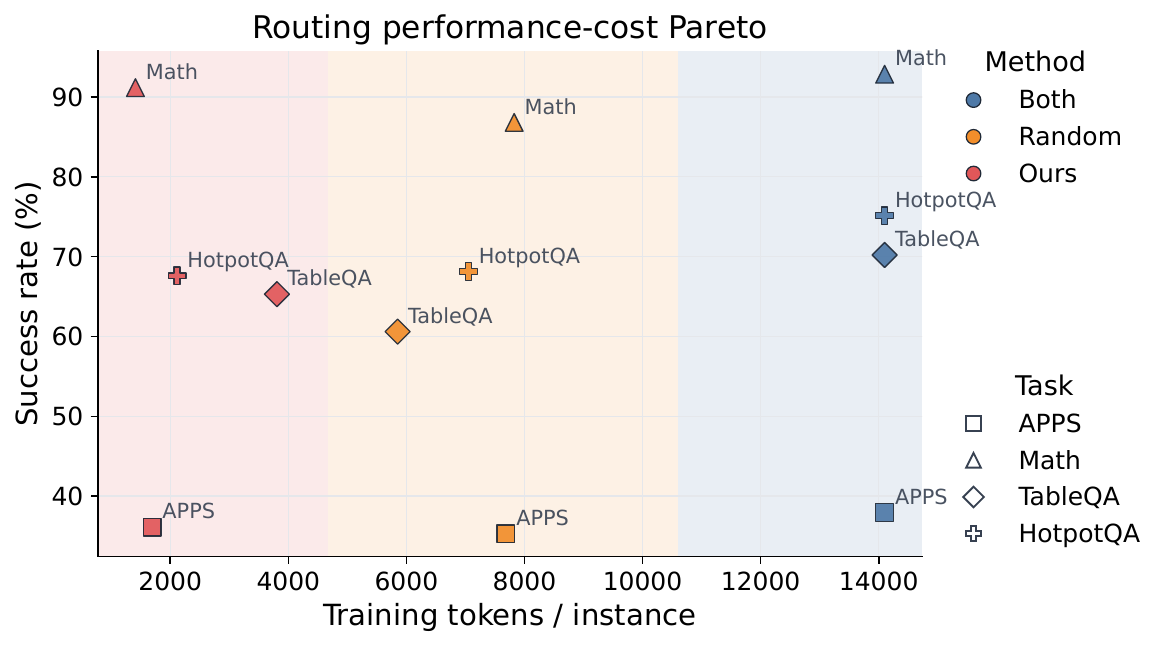}
  \caption{Routing Pareto: COVE preserves most Always-Both performance with far fewer parametric training tokens.}
  \label{fig:router_pareto}
  \vspace{-0.6cm}
\end{figure}

To test the effectiveness of the Task-aware Router, we compare three routing strategies on APPS, MATH, TableQA, and HotpotQA. Lean4 is excluded from this analysis because it requires an additional cold-start stage before routing, which makes the cost comparison less directly comparable across routing strategies. \textbf{Always-Both} uses both channels, representing the most expensive indiscriminate combination. \textbf{Random-Route} randomly assigns instances with equal probability. \textbf{Ours} uses the task-aware router described in Section~\ref{sec:task-aware-routing}. We use the parametric-side token accounting protocol as detailed in Appendix~\ref{app:token-accounting}. Under this protocol, \textit{Token/Inst.} reports the average training cost per routed instance after aggregating over tasks, while \textit{Rel.} and \textit{Eff} summarize relative cost and performance per token.

Table~\ref{tab:router_analysis} and Figure~\ref{fig:router_pareto} show that the router achieves the intended performance-cost trade-off\footnote{We also verified the consistency of routing results implemented based on different backbone LLMs. Tests show that the results obtained using Qwen3-8B, GPT-4o-mini\cite{achiam2023gpt}, DeepSeek-v3.2\cite{liu2024deepseek} achieve over 95\% consistency on these tasks.}. Always-Both obtains the highest average performance, but it pays the full parametric cost for every routed instance. Random-Route cuts the cost roughly in half, yet its performance drops substantially, indicating that simply reducing training frequency is not enough. In contrast, Ours reaches 65.0 average performance while using only 0.16$\times$ the parametric token cost of Always-Both. Its efficiency rises to 28.83, far above both Always-Both and Random-Route. This suggests that the router does not merely save compute, it saves compute mainly on instances for which parametric learning is unlikely to be the most useful update.

Regarding the inter-task routing behavior, contrary to our initial expectations, the Router assigns more MATH tasks to the harness channel. This is because although the MATH dataset contains theorems and knowledge, model already performs well during the inference phase. As a result, feeding these samples into the parametric channel would lead to a waste of resources. TableQA receives the largest parametric-side share among all tasks, because repeated table-operation patterns are reusable across instances. However, the router still keeps many examples on the harness side because column names, schemas, and tool surfaces remain instance-dependent.
\subsection{Anti-Recitation on Volatile API Knowledge}
\begin{figure*}[t]
  \centering
  \includegraphics[width=1\textwidth]{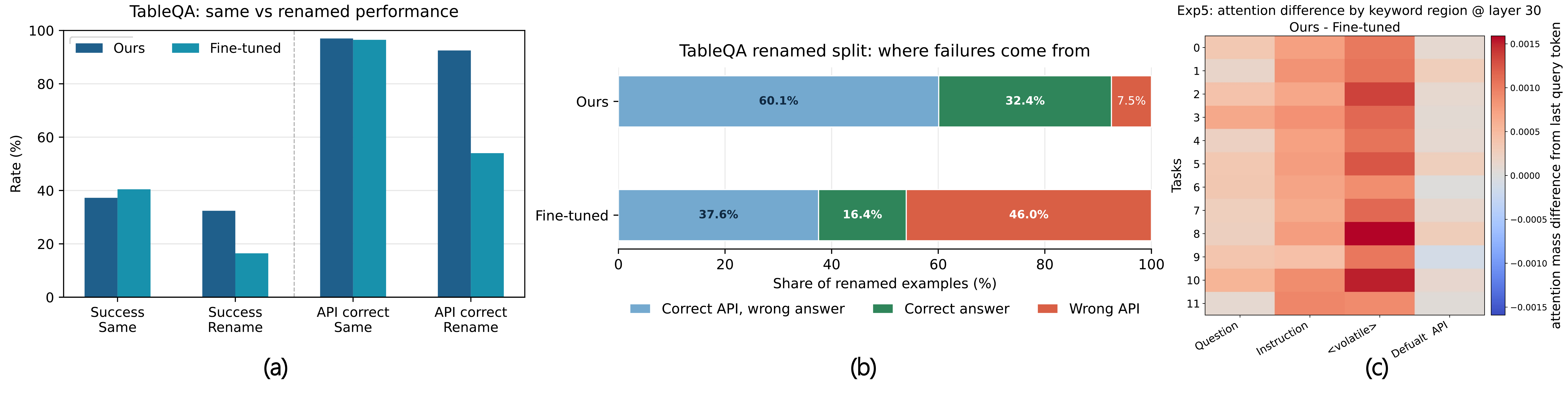}
  \vspace{-0.3cm}
  \caption{API-renaming analysis: anti-recitation preserves API correctness and shifts attention toward current instructions.}
  \label{fig:anti_recitation}
\end{figure*}
\begin{figure*}[t]
  \centering
  \includegraphics[width=1\textwidth]{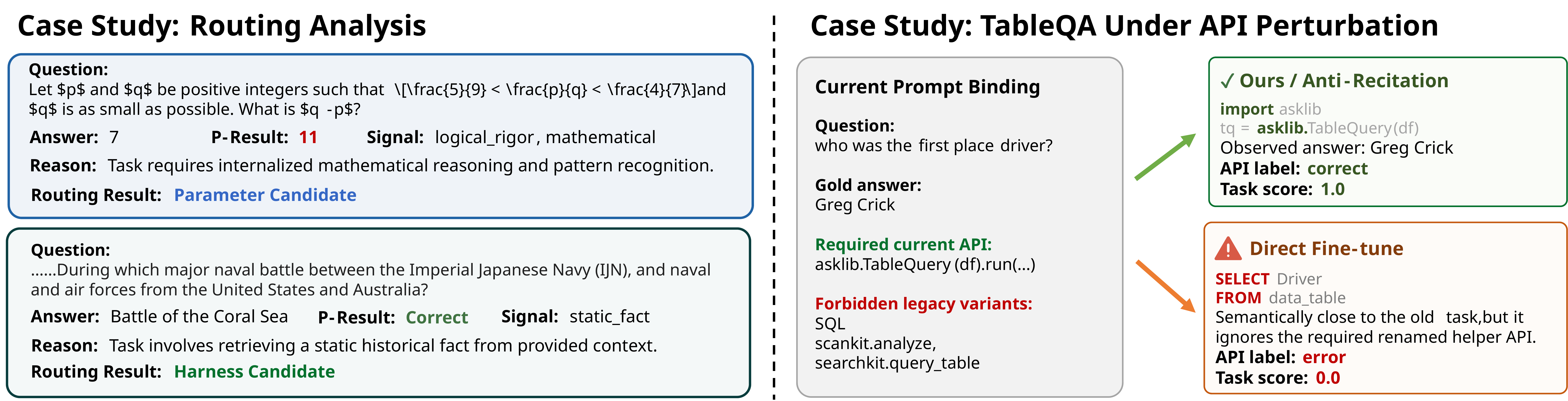}
  \vspace{-0.3cm}
  \caption{Router and API-perturbation cases: COVE separates parametric from harness updates and follows renamed APIs instead of stale interfaces.}
  \label{fig:5_case_study}
  \vspace{-0.3cm}
\end{figure*}

We next test whether volatility-aware knowledge classification and the anti-recitation reward can prevent the model from internalizing volatile interface knowledge. We use the TableQA setting, where successful completion requires explicit API calls. Starting from the same training trajectories, we compare two training variants: a standard fine-tuned model and an anti-recitation model (Ours) that marks volatile API knowledge and applies the anti-recitation reward during training. We then evaluate both models on two test splits. In the \emph{same} split, API names are unchanged from training. In the \emph{rename} split, the API names are deliberately obfuscated, while the task semantics remain unchanged.

Figure~\ref{fig:anti_recitation}(a) shows a clear robustness pattern. On the \emph{same} split, the two models achieve similar success rates and nearly identical API-call correctness, indicating that the anti-recitation objective does not harm in-distribution performance. On the \emph{rename} split, however, the gap becomes substantial: standard fine-tuned model suffers a large drop in both task success and API correctness, whereas the anti-recitation model remains much more stable. This result verifies the intended role of anti-recitation: volatile interface names should remain editable in external memory.

Figure~\ref{fig:anti_recitation}(b) further decomposes the failure modes under API perturbation. When the anti-recitation model fails, its dominant error type is \emph{correct API but wrong answer}. In contrast, fine-tuned model is dominated by \emph{wrong API} errors, showing that its failures often occur before task reasoning begins. This supports the mechanism suggested by the aggregate accuracy: anti-recitation mainly protects the interface-selection step from stale parametric memory.

To further understand this effect, we compare the attention distribution of the two models. Figure~\ref{fig:anti_recitation}(c) reports the attention-mass difference over several keyword regions for the same inputs. Darker red indicates that the anti-recitation model assigns more attention to that region than the fine-tuned model. The strongest increases appear on the \texttt{<volatile>} marker and the instruction region, while the default API region receives much smaller gains. This suggests that the anti-recitation objective helps the model explicitly recognize volatile knowledge and rely more on the current instruction, which contains the renamed API description.

\subsection{Case Study: Routing and API Perturbation}
Figure~\ref{fig:5_case_study} provides qualitative evidence for the two design choices studied above. The routing examples show that the router separates instances according to the kind of knowledge exposed by feedback, rather than by task label alone. The MATH example is routed to the parametric channel because predicting 11 instead of the correct answer 7 indicates a reusable reasoning deficiency involving \texttt{logical\_rigor} and mathematical reasoning patterns. In contrast, the QA example is routed to the harness channel because the model already obtains the correct answer from the provided context, suggesting that the useful knowledge is instance-specific rather than suitable for parameter internalization.

The API-perturbation example illustrates the same distinction for volatile knowledge. Although the renamed TableQA task is semantically close to training examples, the valid tool surface has changed: the current prompt requires \texttt{asklib} and explicitly rules out legacy variants such as SQL. The anti-recitation model follows this current binding and obtains the correct answer, while fine-tuning model still emits the memorized SQL query. This explains why the gap in Figure~\ref{fig:anti_recitation} is concentrated in API correctness: without anti-recitation, the model can learn the old interface as a parametric habit even when the prompt provides the updated API.

\section{Conclusion}

This paper studies how LLM agents should coordinate external memory and parametric learning during self-evolution. We show that neither harness-based nor parameter-based evolution is sufficient on its own. To address this trade-off, we introduce COVE, which routes tasks across evolution channels, schedules parametric updates by stage-aware signals, and separates stable from volatile knowledge through dual-modal optimization. Experiments across reasoning, QA, coding, and theorem-proving tasks show that coordinated evolution improves robustness and efficiency over single-channel alternatives. These results suggest that future self-evolving agents should treat feedback not as uniform training data, but as heterogeneous knowledge whose storage and update mechanism must match its stability and reuse pattern.

\bibliographystyle{ACM-Reference-Format}
\bibliography{sigconf}

\appendix

\section{Token Accounting}
\label{app:token-accounting}

For Table~\ref{tab:exp1_success_rate_cost}, the reported token cost is the number of tokens consumed to train the corresponding model, amortized over the evaluation instances so that it is directly comparable across methods. This cost is modeled in a training-set-independent manner as the product of a routing factor, namely the fraction of instances actually sent through the parametric channel, and a per-task single-step cost anchored to the measured GRPO completion length. The harness memory channel is not charged. Methods that take no parametric step (Base, Evo-Memory, Harness-only), the Lean4 task, and the hybrid split are omitted. For Self-Challenging, a single supervised model serves all tasks, so the same flat cost is charged in every column. 

Although Self-Challenging optimizes the policy with reinforcement learning, its reported self-improvement setting uses the same LLM to generate tasks, collect trajectories, and assign verification-function rewards. Since Self-Challenging uses a binary outcome reward, its one-step REINFORCE objective is equivalent to rejection fine-tuning, i.e., supervised fine-tuning on successful trajectories only. We therefore account for Self-Challenging using the same training-token unit as supervised parametric updates.

\begin{algorithm*}[ht]
\small
\caption{COVE Workflow}
\label{alg:ase-workflow}
\begin{algorithmic}[1]
\REQUIRE Task stream $\{\tau_i\}_{i=1}^{T}$, policy $\pi_\theta$, knowledge base $\mathcal{K}$, environment $\mathcal{E}$
\ENSURE Updated policy $\pi_\theta$ and maintained knowledge base $\mathcal{K}$

\FOR{each incoming task $\tau_i$}
    \STATE Retrieve relevant Memory/Skill entries: $\mathcal{K}_r \gets \textsc{Retrieve}(\mathcal{K}, \tau_i)$
    \STATE Run harness-based inference in $\mathcal{E}$ with $\mathcal{K}_r$
    \STATE Collect feedback $f_i$
    \STATE Extract candidate entries with volatility labels: $\mathcal{C}_i \gets \textsc{ExtractCandidates}(\tau_i, f_i)$

    \FOR{each candidate $c \in \mathcal{C}_i$}
        \STATE $\mathcal{K} \gets \textsc{UpdateKnowledgeBase}(\mathcal{K}, c)$ \COMMENT{insert, merge, revise, discard, and update statistics}
    \ENDFOR
    \STATE Promote repeatedly revised entries to \texttt{volatile}

    \STATE $m_i \gets \textsc{Route}(\tau_i, f_i, \text{history}, \mathcal{K}_r)$
    \COMMENT{\texttt{harness\_only}, \texttt{parametric\_candidate}, or \texttt{hybrid}}

    \WHILE{\NOT \textsc{Trigger}$(\tau_i, f_i, \text{history}, \mathcal{K})$}
        \STATE Continue harness-side exploration with updated $\mathcal{K}$
        \STATE Collect additional feedback and refine retrieved entries
    \ENDWHILE

    \IF{$m_i \in \{\texttt{parametric\_candidate}, \texttt{hybrid}\}$}
        \STATE $\mathcal{D}_{rollout} \gets \textsc{CollectOrReuseRollouts}(\tau_i, \mathcal{K})$
        \STATE Train $\pi_\theta$ on $\mathcal{D}_{rollout}$ with injected Memory/Skill entries
        \STATE Apply task reward and anti-recitation penalty to volatile knowledge
        \STATE $\mathcal{S} \gets \textsc{EvaluateStableMemories}(\mathcal{K})$

        \FOR{each stable entry $k \in \mathcal{S}$}
            \IF{\textsc{UsefulButNowUnnecessary}$(k)$}
                \STATE Mark $k$ as \texttt{internalized}
            \ELSIF{\textsc{NoDemonstratedUtility}$(k)$}
                \STATE Discard $k$
            \ELSE
                \STATE Keep $k$ available for online retrieval
            \ENDIF
        \ENDFOR
    \ENDIF

    \STATE Resume harness-based interaction with updated $\pi_\theta$ and $\mathcal{K}$
\ENDFOR
\end{algorithmic}
\end{algorithm*}
For the routing analysis in Table~\ref{tab:router_analysis}, the accounting objective is different from that of the main experiment. The main table compares end-to-end methods after their own training-data construction procedures, so it amortizes the total model-training cost over evaluation instances. The routing analysis instead isolates the effect of the routing decision itself on training-token use. To avoid confounding this comparison with different numbers of available training samples across strategies, we compute the training token cost per instance. Let $n_{\mathrm{param}}$ denote the number of routed instances whose decision activates the parametric channel for a task. We estimate
\[
\mathrm{train\_tokens}=n_{\mathrm{param}}\times 14100,
\mathrm{train\_steps}=n_{\mathrm{param}}\times 0.125 .
\]
Thus, Always-Both assigns all routed instances to the parametric channel, whereas a selective router pays this cost only for instances judged to contain internalizable training signal. In Table~\ref{tab:router_analysis}, \textit{Token/Inst.} report the average parametric-side training cost per routed instance after aggregating over tasks. \textit{Rel.} is the average token cost of the method divided by the average token cost of Always-Both. \textit{Eff} is computed as the average performance percentage divided by the average token cost and then multiplied by one thousand; it is therefore a performance-per-token summary, not an average of per-instance efficiencies.

\section{COVE Workflow}
\label{app:cove-workflow}
The workflow of COVE is presented in Algorithm~\ref{alg:ase-workflow}.

\end{document}